\documentclass{article}
\usepackage{iclr2026_conference,times}

\usepackage{amsmath,amsfonts,bm}

\def\eqref#1{equation~\ref{#1}}

\def\1{\bm{1}}

\DeclareMathAlphabet{\mathsfit}{\encodingdefault}{\sfdefault}{m}{sl}
\SetMathAlphabet{\mathsfit}{bold}{\encodingdefault}{\sfdefault}{bx}{n}

\usepackage{color}
\usepackage{multirow}
\usepackage{algorithmicx,algorithm}
\usepackage[noend]{algpseudocode}
\usepackage{subfigure}
\usepackage{caption}
\usepackage{enumitem} 
\usepackage{xspace}
\usepackage{latexsym}
\usepackage{amsmath}
\usepackage{booktabs}
\usepackage{tabularx}
\usepackage{graphicx}
\usepackage{svg}
\usepackage{listings}
\usepackage[most]{tcolorbox}
\usepackage{pifont}
\svgsetup{
    inkscapepath=i/svg-inkscape/
}
\svgpath{{svg/}}
\usepackage{amssymb}
\usepackage{etoolbox}
\usepackage[all]{nowidow}
\usepackage[normalem]{ulem}
\usepackage{hyperref}
\hypersetup{
  hidelinks
}

\usepackage{url}

\setlist{nolistsep}
\newcommand{\parabf}[1]{\medskip\noindent\textbf{#1}}
\newcommand{\parait}[1]{\medskip\noindent\textit{#1}}
\newcommand{\paraf}[1]{\noindent\textbf{#1}}
\newcommand{\cut}[1]{}

\newcommand{\sysname}{\textsc{BigMac}\xspace}

\definecolor{codeblue}{RGB}{34,101,150}
\definecolor{codered}{RGB}{190,35,55}
\definecolor{codegreen}{RGB}{68,128,50}
\definecolor{refgreen}{RGB}{0,204,0}
\definecolor{codegray}{RGB}{128,128,128}
\definecolor{codebg}{RGB}{249,249,247}
\definecolor{codepurple}{RGB}{105,70,160}
\lstdefinestyle{bigmacpy}{
    language=Python,
    basicstyle=\ttfamily\scriptsize,
    keywordstyle=\color{codered}\bfseries,
    commentstyle=\color{codegray}\bfseries,
    stringstyle=\color{codeblue},
    identifierstyle=\color{black},
    alsoletter={_},
    morekeywords={def,return,if,for,while,in},
    emph={def,return,if,for,while,in},
    emphstyle=\color{codered}\bfseries,
    emph={[2]build_schedule,encoder_forward,llm_forward_step,generator_forward,run_bigmac,partition_microbatches,columns,unit,llm_output_ready,encoder_grad_ready,get_llm_schedule,insert_comm_ops,deadlock_check,create_executor,len,min,next},
    emphstyle={[2]\color{codeblue}\bfseries},
    emph={[3]enc_fwd,enc_bwd,gen_fwd,gen_bwd,encoder,llm,generator,embed_preprocess,load_schedule,execute},
    emphstyle={[3]\color{codegreen}\bfseries},
    emph={[4]llm_schedule,llm_iter,data_iter,encoder_inputs,input_ids,encoder_feature,embedding,output,loss,executor,units,schedule,next_unit,column,op,u,unit_size,pp_size,microbatch_num,warmup_bound},
    emphstyle={[4]\color{codepurple}},
    columns=fullflexible,
    keepspaces=true,
    showstringspaces=false,
    frame=none,
    tabsize=2,
    breaklines=true,
    escapeinside={(*@}{@*)}
}

\title{\sysname: Breaking the Pareto Frontier of Compute and Memory in \\ Multimodal LLM Training}

\author{
Zili Zhang$^1$\thanks{Equal contribution.}\hspace{0.45em}\footnotemark[2]\hspace{0.35em}, Chengxu Yang$^2$\footnotemark[1] , Shenglong Zhang$^2$, Chenyu Wang$^2$, Yufan Zhang$^2$,\\
\bfseries Tuo Dai$^3$, Zhouyang Li$^2$, Yuhong Ge$^3$, Chao Jin$^1$\thanks{Work done during internships at Xiaohongshu, Inc..} , Xin Jin$^1$\thanks{Co-corresponding authors.} , Yuliang Liu$^2$\footnotemark[3]\\
\normalfont $^1$Peking University \qquad $^2$Independent Researcher \qquad $^3$Xiaohongshu, Inc.
} 

\iclrfinalcopy 

\begin{document}
\sloppy

\maketitle

\begin{abstract}
Training multimodal large language models (MLLMs) is challenged by both \textbf{model} and \textbf{data heterogeneity}.
Existing systems redesign the training pipeline to address these challenges, but remain bound by a Pareto
frontier between compute and memory efficiency, improving one only at the expense of the other.
We present \sysname, a new training pipeline for multimodal LLMs. The core idea of \sysname is to elegantly nest the encoder
and generator computation into the
original LLM pipeline, forming a \textbf{dependency-safe nested pipeline} structure. With this design,
\sysname reduces the activation memory complexity of the encoder and generator to $O(1)$ while keeping the activation memory complexity of
the LLM unchanged. At the same time, it achieves the same computational efficiency as the idealized setting with unlimited memory.
As a result, \sysname breaks the Pareto frontier between computational efficiency and memory usage, enabling simultaneous optimization of both computation and memory in MLLM training.
We evaluate \sysname on multiple MLLMs and training workloads. Experimental results show that \sysname achieves
a 1.08$\times$--1.9$\times$ training speedup over baseline systems while maintaining stable memory usage as batch size increases.
\end{abstract}

\begin{figure}[h!]
    \centering
    \includegraphics[width=0.8\linewidth]{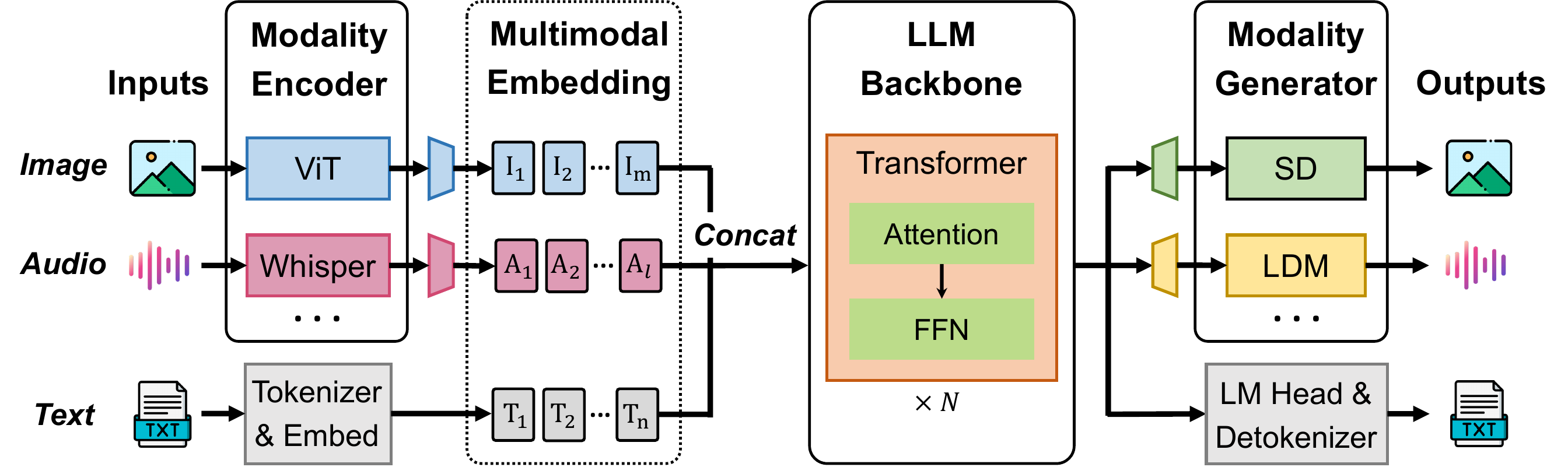}
    \vspace{-0.1in}
    \caption{The architecture of multimodal LLMs.}
    \vspace{-0.17in}
    \label{fig:intro:model}
\end{figure}

\section{Introduction}
\label{sec:introduction}

Multimodal large language models (MLLMs) provide a unified framework for processing multimodal inputs and generating
multimodal outputs, making them a powerful foundation for next-generation AI systems. By combining the reasoning,
instruction-following, and generative capabilities of LLMs with modality-specific components, MLLMs enable more
general and flexible interaction across modalities such as vision, audio, and language. This capability has made
MLLMs increasingly important for a broad range of applications, including visual understanding~\citep{liu2023visual} and generation~\citep{chameleon2024mixed}, audio
understanding~\citep{chu2024qwen2audio}, embodied AI~\citep{driess2023palme,brohan2023rt2}, and interactive multimodal assistants~\citep{openai2023gpt4v}. The rapid progress of recent MLLM systems further
demonstrates their growing importance in both research and real-world deployment.

Figure~\ref{fig:intro:model} illustrates a representative multimodal LLM architecture~\citep{openai2023gpt4v,wu2025qwen}, consisting of a modality encoder, an LLM backbone, and a modality generator connected by projectors such as MLPs or cross-attention modules.
The encoder maps multimodal inputs into modality tokens, which are organized into a token sequence and processed by the LLM; the generator maps the LLM outputs back to the target modality.
Training such architectures is challenging due to both model and data heterogeneity, as identified by DistTrain~\citep{zhang2025disttrain}.
Model heterogeneity arises from the structural and computational differences among the encoder, LLM backbone, and generator, which exhibit distinct execution patterns.
Data heterogeneity further complicates training: even with a fixed sequence length, multimodal samples may contain different numbers of modality tokens, causing workload imbalance and inefficient resource utilization.

Existing systems address these challenges by redesigning the training pipeline, but remain constrained by a Pareto frontier between compute and memory efficiency.
Compute-efficient systems, such as LoongCat~\citep{team2025longcat} and Optimus~\citep{feng2025optimus}, improve throughput by decoupling encoder computation from the LLM pipeline and executing encoder outside the LLM pipeline.
While this eliminates computational interference between the encoder and LLM, it requires retaining all encoder activations until the LLM pipeline finishes, resulting in $O(M/P)$ encoder activation memory, where $M$ is the number of microbatches and $P$ is the pipeline parallelism degree.
With a generator, the problem is further amplified because LLM activations must also be preserved until generator computation completes.
In contrast, memory-efficient systems, such as DistTrain~\citep{zhang2025disttrain} and Megatron~\citep{shoeybi2019megatron}, integrate the encoder and generator into the LLM pipeline as its initial and final layers.
This reduces the activation memory complexity of the encoder and generator to $O(P)$ and $O(1)$, respectively.
However, sequentially executing all components in one pipeline introduces computational interference, making the iteration time dictated by the slowest stage.
Consequently, these systems achieve substantially lower compute efficiency than compute-efficient alternatives.

To this end, we present \sysname, a new training framework for multimodal LLMs.
Its key idea is a dependency-safe nested pipeline that interleaves the encoder and generator with the original LLM pipeline without altering its dependencies.
This design matches the compute efficiency of an idealized compute-efficient setting with unlimited memory, while reducing encoder and generator activation memory to $O(1)$.
As a result, \sysname breaks the Pareto frontier between compute efficiency and memory usage, enabling both high compute efficiency and low memory usage in MLLM training.
\section{Background and Related Work}
\label{sec:background}

\subsection{Background}
\paraf{Multimodal LLMs.}
Large language models (LLMs)~\citep{openai2023gpt4v,wu2025qwen} have demonstrated strong capabilities in language understanding and generation, but their native interface is limited to text.
Multimodal LLMs (MLLMs) extend LLMs with modality-specific components to process multimodal inputs such as images and audio and generate outputs beyond plain text.
As shown in Figure~\ref{fig:intro:model}, a representative MLLM consists of modality encoders, an LLM backbone, and modality generators.
The encoders convert raw modality inputs into intermediate features, e.g., using ViT~\citep{dosovitskiy2021image} for images or Whisper~\citep{radford2023robust} for audio, which are then projected into the LLM embedding space through MLP or cross-attention projectors.
The LLM backbone performs multimodal reasoning over the resulting embeddings, and the output representations are mapped by output projectors to modality generators, such as diffusion~\citep{rombach2022high} for images or AudioLDM~\citep{liu2023audioldm} for audio.
Training such architectures is challenging due to model and data heterogeneity, as identified by DistTrain~\citep{zhang2025disttrain}.

\begin{figure*}[t!]
    \centering
    \includegraphics[width=\linewidth]{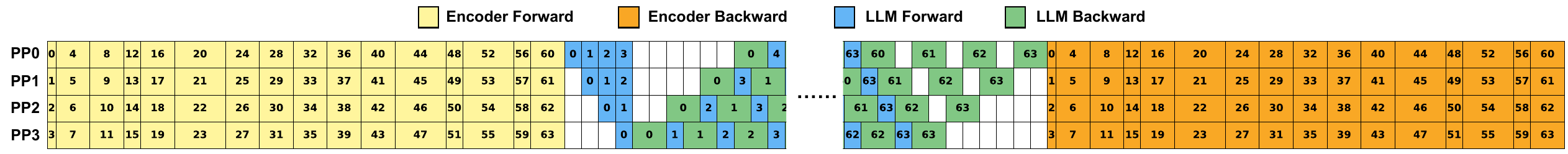}
    \vspace{-0.2in}
    \caption{The pipeline structure of compute-efficient MLLM training (pp size = 4, batch size = 64).}
    \vspace{-0.17in}
    \label{fig:compute_efficient}
\end{figure*}

\parabf{Model heterogeneity.}
MLLMs exhibit substantial model heterogeneity because their constituent modules differ in architecture and computational characteristics.
Modality encoders such as ViT typically use relatively narrow transformer layers, the LLM backbone uses much wider transformer layers, and modality generators such as diffusion models combine convolution and attention operators.
As a result, the encoder, backbone, and generator have distinct execution times, creating imbalance in pipeline-parallel (PP) training.
If an LLM PP stage is the bottleneck, the encoder and generator stages wait, reducing resource utilization.
Conversely, if the encoder or generator is slower than the LLM stages, it stalls the LLM pipeline and introduces bubbles.
This latter case is particularly harmful at scale, where most GPUs are allocated to the LLM backbone, leading to substantial resource underutilization.

\parabf{Data heterogeneity.}
MLLM training data exhibit substantial data heterogeneity because modality subsequences within each sample are highly skewed in both length and count, causing large variation in the number of modality tokens~\citep{deng2025emerging}.
For example, one sample may contain only a few tokens from a single image, while another may contain hundreds or thousands tokens from multiple high-resolution images.
This heterogeneity leads to significant execution-time variation in the modality encoder and generator, even when the LLM backbone has stable execution time under a fixed sequence length.
As a result, pipeline-parallel training suffers from microbatch stragglers: slower microbatches delay faster ones and introduce pipeline bubbles that disrupt pipeline progress~\citep{zhang2025disttrain}.
This further amplifies computation imbalance, reduces GPU utilization, and prolongs end-to-end MLLM training.

\subsection{Related Work}
\paraf{Compute-efficient systems.}
To address model and data heterogeneity, existing compute-efficient systems~\citep{team2025longcat,feng2025optimus} redesign the MLLM training pipeline to eliminate computational interference among the encoder, generator, and LLM backbone.
They typically apply different parallelization strategies to different modules: data parallelism for the encoder and generator, and pipeline parallelism for the LLM backbone.
As shown in Figure~\ref{fig:compute_efficient}, these systems first execute the encoder forward pass for all microbatches and retain the resulting activations for backward computation.
After the encoder forward finishes, the LLM pipeline processes the modality embeddings using a 1F1B schedule~\citep{narayanan2019pipedream}.
By separating encoder and LLM computation along the time dimension, this design avoids cross-module interference; thus, differences in encoder, generator, and LLM execution times do not introduce pipeline bubbles.
However, this compute efficiency comes at the cost of high memory usage.
All encoder activations must be retained until the LLM pipeline completes, and the problem becomes more severe when a generator is included.
In that case, LLM activations must also be preserved until generation finishes.
Since LLM activations are substantially larger than encoder activations, this approach becomes impractical for production-scale MLLM training with modality generators.

\begin{figure*}[t!]
    \centering
    \includegraphics[width=\linewidth]{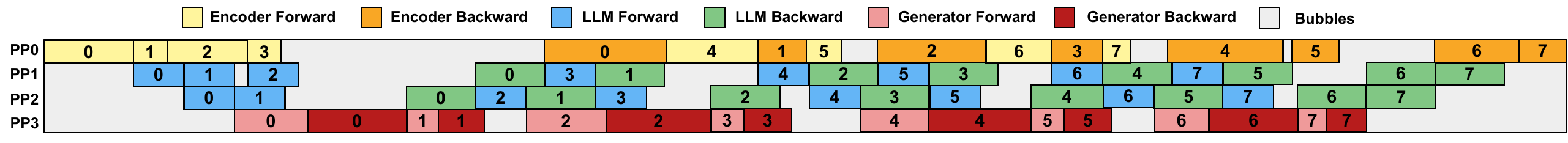}
    \vspace{-0.2in}
    \caption{The pipeline structure of memory-efficient MLLM training (pp size = 4, batch size = 8).}
    \vspace{-0.17in}
    \label{fig:memory_efficient}
\end{figure*}

\parabf{Memory-efficient systems.}
Memory-efficient systems~\citep{zhang2025disttrain, shoeybi2019megatron} take the opposite approach by integrating the modality encoder and generator into the LLM pipeline as the initial and final layers, respectively.
As shown in Figure~\ref{fig:memory_efficient}, they execute the encoder, LLM backbone, and generator sequentially in a single 1F1B pipeline.
This design reduces the activation memory complexity of the encoder and generator to $O(P)$ and $O(1)$, respectively, where $P$ is the pipeline-parallelism degree.
However, this memory efficiency comes at the cost of compute efficiency.
While LLM microbatches have relatively stable execution time under a fixed sequence length, encoder and generator microbatches can vary substantially due to model and data heterogeneity.
Because all components are coupled by sequential pipeline dependencies, this variation introduces bubbles across the pipeline, as shown in Figure~\ref{fig:memory_efficient}.
Although automatic parallelism tuning~\citep{zheng2022alpa} and dynamic microbatch scheduling~\citep{zhang2025disttrain} can partially mitigate the issue, they still fall short of the compute efficiency achieved by compute-efficient systems.

\parabf{Performance analysis.}
The two baselines above expose a fundamental memory--compute tradeoff.
Let $M$ be the number of microbatches and $P$ the LLM pipeline-parallelism degree, and assume that one encoder/generator microbatch has activation footprint $A_m$ while one full LLM microbatch has footprint $A_l$.
Compute-efficient systems decouple encoder and generator computation from the LLM pipeline, so the LLM only incurs the standard 1F1B fill/drain overhead, with bubble rate approximately $(P-1)/(M+P-1)$, which becomes small when $M \gg P$.
However, this low bubble rate comes from extending activation lifetime: on the first pipeline stage, peak activation memory is approximately $\frac{M}{P}A_m + A_l$, because modality activations for all assigned microbatches must be retained while the LLM pipeline executes.
Memory-efficient systems instead integrate the encoder and generator into the pipeline, bounding peak modality activation memory by $P A_m$ on the first PP stage or $\frac{P-1}{P}A_l$ on the second PP stage, thereby removing the memory term that grows with $M$.
However, this design introduces additional steady-state bubbles beyond 1F1B fill/drain overhead, because encoder and generator microbatches may have different execution times.
Thus, when $M$ is large, compute-efficient systems achieve low bubble rates at high activation-memory cost, while memory-efficient systems reduce memory by sacrificing compute efficiency.
\section{\sysname Design}
\label{sec:design}

\sysname consists of two main components: the \sysname pipeline scheduler and the \sysname pipeline executor.
The pipeline scheduler is responsible for determining the optimal scheduling of
microbatches across different modules, while the pipeline executor manages the execution of the
scheduled microbatches on the underlying training framework.

\begin{figure*}[t!]
    \centering
    \includegraphics[width=\linewidth]{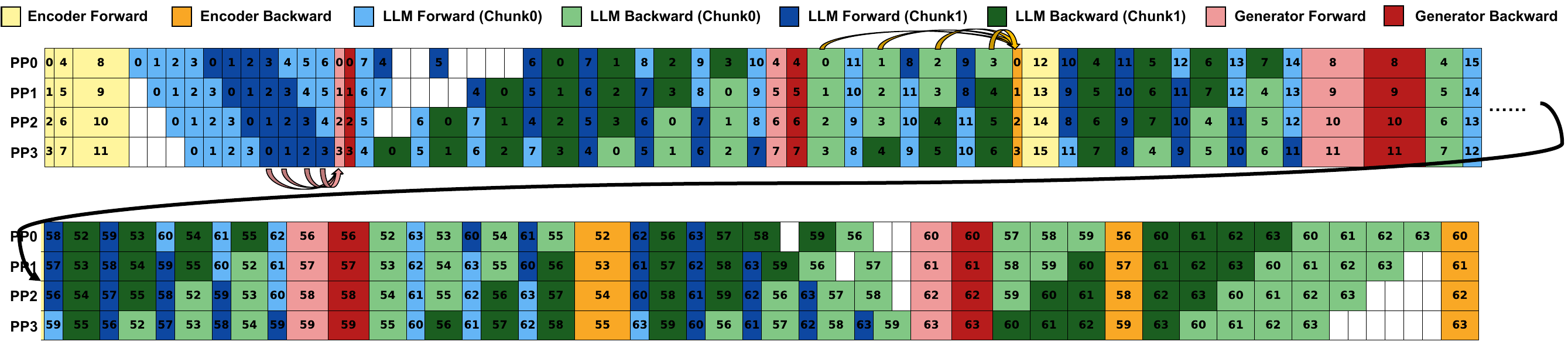}
    \vspace{-0.2in}
    \caption{The pipeline scheme of \sysname (pp size = 4, vpp size = 2, batch size = 64).}
    \vspace{-0.17in}
    \label{fig:bigmac}
\end{figure*}

\subsection{\sysname Pipeline Scheduler}
In this section, we present the details of \sysname's pipeline scheduler, which is responsible
for orchestrating the execution of \uline{compute} and \uline{communication} operators
by \textbf{dependency-safe nesting}.

\parabf{Compute operator.}
Figure~\ref{fig:bigmac} shows the overall pipeline scheme.
The scheduler targets two goals: preserving the compute efficiency of the LLM pipeline and bounding encoder/generator activation memory.
To avoid computational interference, \sysname treats the encoder, LLM, and generator as separate parallelism domains rather than a single stage-by-stage pipeline.
This matches production MLLM training, where the LLM backbone usually requires pipeline parallelism, while smaller encoders and generators can run with DDP~\citep{li2020pytorchdistributed} or FSDP~\citep{zhao2023pytorchfsdp}.
Thus, within one pipeline group, an LLM micro-batch traverses all pipeline ranks, whereas encoder and generator micro-batches are processed locally.

\parait{\underline{Compute efficiency.}}
\sysname groups encoder and generator operators into units of $pp\_size$ micro-batches, where $pp\_size$ is the LLM pipeline parallelism size.
Each unit corresponds to one LLM pipeline column in Figure~\ref{fig:bigmac}.
The scheduler starts from the user-specified LLM schedule, such as interleaved 1F1B, and nests encoder and generator units within it.
Since encoder and generator units are not part of the LLM pipeline stages,
\sysname can nest encoder/generator units into the LLM schedule without introducing bubbles.
Furthermore, this nested structure
preserves the original LLM dependencies while eliminating cross-module execution interference.

\parait{\underline{Memory efficiency.}}
For memory efficiency, \sysname executes encoder and generator backward operators as soon as their dependencies are satisfied.
This requires respecting cross-module edges: LLM forward consumes encoder outputs, and LLM backward depends on gradients produced through generator backward.
\sysname therefore places each encoder and generator unit at the earliest dependency-safe point without changing the LLM order.
It first warms up a constant number ($W$) of encoder forward units, then pairs each ready encoder backward unit with the next encoder forward unit.
For the generator, it runs forward as soon as LLM outputs are ready and backward immediately afterward.
This policy preserves all cross-module dependencies, keeps the LLM schedule unchanged, and requires only $W=3$ encoder activations and one generator activation.
Figure~\ref{fig:build_schedule} shows the pseudocode of the scheduling algorithm.

\parait{\underline{Analysis.}}
For interleaved 1F1B schedules with $vpp\_size \ge 2$, \sysname sets $W=3$, because such schedules require only a three-unit encoder lookahead.
Let $U_i$ denote the $i$-th encoder unit, which contains $pp\_size$ consecutive micro-batches.
Let $F_i$ denote the first entry-side LLM forward operator that consumes the encoder output of $U_i$, and let $G_i$
denote the point at which the LLM produces all the input gradient required by $U_i$'s backward.
At the unit level, interleaved 1F1B with $vpp\_size \ge 2$ satisfies the following order property:
\[
    F_i,\; F_{i+1},\; F_{i+2} \;\prec\; G_i \;\prec\; F_{i+3}.
\]
That is, after the LLM finishes consuming the output of $U_i$, it consumes the outputs of $U_{i+1}$ and $U_{i+2}$ before the gradient for $U_i$ returns, and this gradient becomes ready before the LLM needs the output of $U_{i+3}$.
This property follows from the steady phase of 1F1B: each new forward wave is matched by the backward wave of an older unit, so the distance between forward consumption and backward return is bounded by three encoder units.
Therefore, before $F_{i+3}$ is reached, \sysname can execute the backward of $U_i$ to release the $U_i$'s activations, and then execute the forward of $U_{i+3}$ to prepare the next unit.
As a result, the ready window slides from $U_i,U_{i+1},U_{i+2}$ to $U_{i+1},U_{i+2},U_{i+3}$ before the LLM consumes $U_{i+3}$.
This invariant ensures that only three encoder-unit activations need to be kept in memory to keep the LLM pipeline fully fed, yielding $O(1)$ encoder activation memory.
By induction on $i$, encoder outputs are always ready when needed, so encoder units neither stall nor reorder the LLM pipeline, allowing \sysname to match the compute-efficient pipeline time.

\begin{figure*}[t]
\centering
\begin{minipage}[t]{0.48\textwidth}
\begin{tcolorbox}[
    enhanced,
    colback=codebg,
    colframe=black!12,
    boxrule=0.4pt,
    arc=1pt,
    left=2mm,
    right=2mm,
    top=1mm,
    bottom=1mm]
\begin{lstlisting}[style=bigmacpy,basicstyle=\ttfamily\tiny]
# M: micro-batches; P: pipeline size
# W: warmup number of encoder units
def build_schedule(llm_schedule, M, P, W):
    units = partition_microbatches(M, unit_size=P)
    schedule, next_unit = [], 0

    # Warm up encoder units.
    while next_unit < min(W, len(units)):
        schedule += [enc_fwd(units[next_unit])]
        next_unit += 1

    # Preserve the original LLM pipeline order.
    for column in columns(llm_schedule):
        schedule += column

        for op in column:
            u = unit(op)
            if llm_output_ready(op, u):
                schedule += [gen_fwd(u), gen_bwd(u)]

            if encoder_grad_ready(op, u):
                schedule += [enc_bwd(u)]
                if next_unit < len(units):
                    schedule += [enc_fwd(units[next_unit])]
                    next_unit += 1

    return schedule
\end{lstlisting}
\end{tcolorbox}
\caption{\textsc{BuildSchedule} nests encoder/generator units into the LLM pipeline schedule.}
\label{fig:build_schedule}
\end{minipage}
\hfill
\begin{minipage}[t]{0.48\textwidth}
\begin{tcolorbox}[
    enhanced,
    colback=codebg,
    colframe=black!12,
    boxrule=0.4pt,
    arc=1pt,
    left=2mm,
    right=2mm,
    top=1mm,
    bottom=1mm]
\begin{lstlisting}[style=bigmacpy,basicstyle=\ttfamily\tiny]
def encoder_forward(data_iter):
    input_ids, encoder_inputs = next(data_iter)
    encoder_feature = encoder(encoder_inputs)
    return input_ids, encoder_feature

def llm_forward_step(llm_iter):
    input_ids, encoder_feature = next(llm_iter)
    embedding = embed_preprocess(input_ids,
                                 encoder_feature)
    output = llm(embedding)
    return output

def generator_forward(output):
    loss = generator(output)
    return loss

def train_step(data_iter, pp_size,
               microbatch_num, warmup_bound):
    # E.g., interleaved 1F1B schedule
    llm_schedule = get_llm_schedule(pp_size,
                                    microbatch_num)
    # BigMac pipeline schedule
    schedule = build_schedule(llm_schedule,
                              microbatch_num,
                              pp_size,
                              warmup_bound)
    schedule = insert_comm_ops(schedule)
    schedule = deadlock_check(schedule)
    executor = create_executor(data_iter,
                               llm_forward_step,
                               encoder_forward,
                               generator_forward)
    executor.load_schedule(schedule)
    executor.execute()
\end{lstlisting}
\end{tcolorbox}
\caption{\sysname pipeline executor interface.}
\label{fig:executor}
\end{minipage}
\end{figure*}

\parabf{Communication operator.}
\sysname makes the compute schedule executable by materializing its implicit data dependencies as communication operators.
For each compute operator, \sysname derives a pre-communication operator that fetches required inputs and a post-communication operator that forwards produced tensors to consumers.
This preserves the pipeline compute order in Figure~\ref{fig:build_schedule} while exposing all cross-device activation and gradient transfers to the executor.

\parait{\underline{Pipeline communication.}}
For the LLM backbone, \sysname uses standard point-to-point pipeline communication.
Stage $i$ receives forward activations from stage $i-1$ and sends outputs to stage $i+1$, with symmetric communication in the backward pass.
Under interleaved 1F1B, each communication operator carries the micro-batch and virtual chunk metadata of its compute operator, allowing the executor to distinguish tensors across chunks and micro-batches.

\parait{\underline{Cross-module communication.}}
Cross-module communication connects separately scheduled encoder/generator units with the LLM pipeline.
Encoder forward gathers modality embeddings to the LLM entry stage, while LLM backward scatters input-embedding gradients back to the owning encoder ranks.
The generator handoff is analogous on the LLM output side: the last LLM stage scatters generator inputs and gathers generator-input gradients back.
These operators provide the concrete producer-consumer links among encoder, LLM, and generator compute.


\subsection{\sysname Pipeline Executor}
\sysname's pipeline executor turns the operator schedule produced by the scheduler into concrete training execution.
Each pipeline rank receives its local schedule, which contains both compute and communication operators.
The executor interprets this schedule as an opcode stream: LLM forward and backward operators are dispatched to user-provided LLM runtime, point-to-point communication operators are dispatched to the runtime communication backend, and encoder/generator operators are dispatched to user-provided module functions.
This design keeps the scheduler independent of framework-specific execution details, while allowing the executor to reuse optimized runtime components such as fused send-receive primitives and DP gradient overlap.

Figure~\ref{fig:executor} illustrates the programming interface of the pipeline executor for executing a training step under a given pipeline schedule.
From the user's perspective, using the executor requires only the forward implementations of the encoder, LLM, and generator.
The executor hides the underlying details of parallelism and communication, providing a simple and easy-to-use interface.

\parabf{Runtime buffers.}
The executor maintains runtime buffers keyed by microbatch id, with an additional virtual chunk id under interleaved 1F1B.
Forward operators write output activations to these buffers, backward operators consume the corresponding activations and produce gradients. Communication operators transfer tensors in the buffer across adjacent pipeline stages.
The executor issues the communication operations and records the returned asynchronous communication handles.
The executor waits on receive handles before consuming received tensors.
Thus, the executor preserves the scheduler-determined operator order while overlapping communication with computation.

\begin{figure*}[t!]
    \centering
    \includegraphics[width=0.9\linewidth]{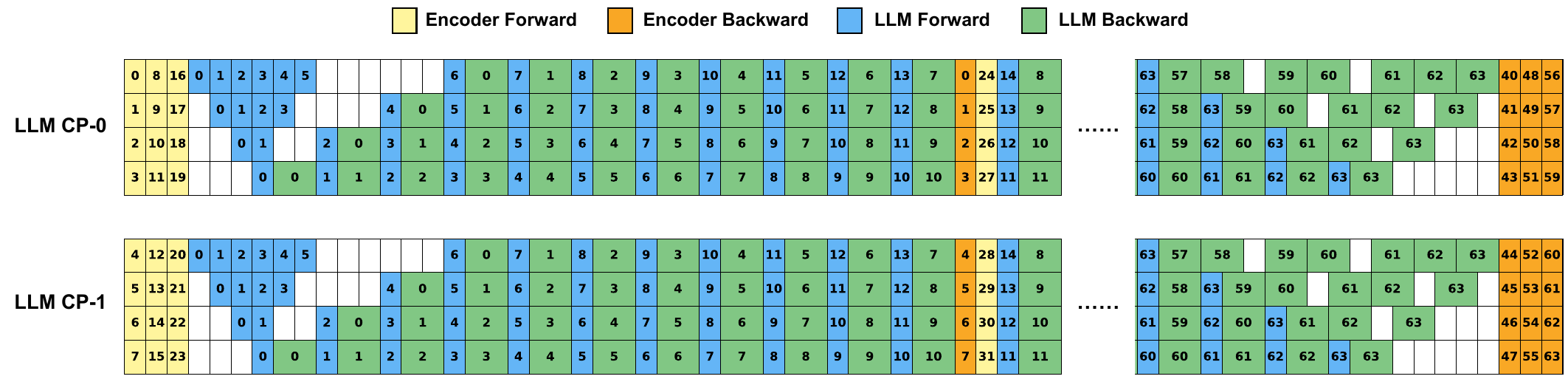}
    \vspace{-0.06in}
    \caption{The pipeline scheme of \sysname with context parallel (LLM CP = 2, Encoder CP = 1).}
    \vspace{-0.17in}
    \label{fig:bigmac_cp}
\end{figure*}

\parabf{Module handoff.}
The executor bridges encoder/generator modules and the LLM through dedicated handoff adapters.
For encoder-to-LLM handoff, it validates encoder outputs using explicit metadata, and then gathers them into the LLM entry-stage.
The communication is launched asynchronously with readiness events so encoder or LLM computation can overlap with gather communication.
The backward path reverses this route: the first LLM stage retains input tensors, collects their gradients after LLM backward, scatters them to the owning encoder ranks, immediately runs encoder backward once a unit's gradient is ready, and releases saved encoder activations.
For distributed data-parallel encoder/generator modules, the executor suppresses non-final gradient synchronization and finalizes gradients at schedule end to preserve standard accumulation semantics.
Generator handoff follows the same pattern on the LLM output side: the last LLM stage scatters output activations to generator ranks, runs generator forward and backward for the unit, and gathers the resulting input gradients to trigger LLM backward.

\subsection{\sysname Compatibility}
In this subsection, we discuss how \sysname integrates with common MLLM training techniques.
We first discuss its compatibility with context parallelism in the LLM backbone for long-context MLLM training.
We then describe its integration with fully sharded data parallelism (FSDP), which is widely used to train modality encoders and generators.

\parabf{Context parallelism}
Context parallelism (CP)~\citep{liu2024ringattention} is a standard technique for training LLMs with long input contexts.
CP partitions the input sequence across multiple ranks and parallelizes attention computation.
In MLLM training, CP may be applied simultaneously to the LLM backbone, encoder, and generator, with different CP degrees for different
modules due to their differing sequence lengths and attention costs.
\sysname supports this multi-module CP setting by allowing the encoder and LLM to use distinct CP groups and by extending the module handoff and pipeline scheduler accordingly.

When the encoder and LLM use different CP groups, the encoder-to-LLM handoff converts modality embeddings from the encoder CP layout to the LLM CP layout through an additional all-to-all communication.
\sysname represents this conversion as an asynchronous CP-conversion operator between encoder operators and LLM operators, using readiness events to overlap communication with computation.
Different CP degrees also constrain pipeline scheduling: pipeline groups within the same LLM CP group consume the same encoder outputs in lockstep because context parallel synchronizes LLM ranks after each micro-batch.
Thus, \sysname enlarges the encoder scheduling unit from $pp\_size$ to $pp\_size \times llm\_cp\_size / encoder\_cp\_size$ micro-batches; in Figure~\ref{fig:bigmac_cp}, for example, LLM CP size two and encoder CP size one yield an eight micro-batch encoder unit while preserving the original LLM pipeline order.

\parabf{Fully sharded data parallelism.}
FSDP is a memory-efficient data-parallel runtime that shards model parameters, gradients, and optimizer states across GPUs.
Unlike standard data parallelism, FSDP only maintains a parameter shard for each data parallel rank during training.
Before executing an FSDP-wrapped module in either the forward or backward pass, all ranks in the FSDP group must first \texttt{all-gather} the full parameters.
As a result, each FSDP module introduces a synchronization point across the group, which may interact poorly with \sysname's nested pipeline scheme.

\begin{figure}[t!]
    \centering
    \includegraphics[width=\linewidth]{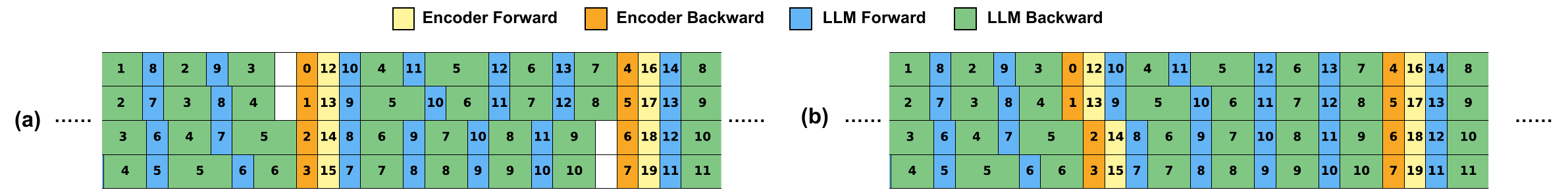}
    \vspace{-0.17in}
    \caption{A pipeline slice of \sysname with an encoder trained using FSDP.}
    \vspace{-0.12in}
    \label{fig:design:fsdp}
\end{figure}

As shown in Figure~\ref{fig:design:fsdp}(a), LLM operator time varies across micro-batches; for example, the backward operator of micro-batch 5 takes longer than the others.
This variation comes from sequence packing, which creates different attention workloads across micro-batches~\citep{ge2025bytescale}, and MoE expert parallelism, where expert load imbalance creates uneven GPU workloads~\citep{deepseek2025eplb}.
Under such non-uniform execution, FSDP's strict synchronization introduces additional pipeline bubbles in the steady phase.

\parait{\underline{One-Sided pull.}}
To solve this issue, \sysname replaces the collective communication, i.e., \texttt{all-gather} of FSDP-wrapped encoder and generator with a request-driven one-sided pull mechanism.
Instead of requiring all ranks in an FSDP group to enter the same \texttt{all-gather} at the same time, each rank publishes its local parameter shard in NVSHMEM-accessible memory.
When a rank reaches a FSDP-wrapped encoder or generator operator, it materializes the required full parameter bucket by fetching the missing shards from their owner ranks using one-sided NVSHMEM \texttt{get} operations.
Thus, \sysname removes the FSDP synchronization barrier, allowing ranks to proceed independently according to the pipeline schedule.
As shown in Figure~\ref{fig:design:fsdp}(b), this design keeps FSDP's sharded-memory benefit while removing the all-rank \texttt{all-gather} barrier that creates the pipeline bubbles.

\section{Evaluation}
\label{sec:evaluation}

\subsection{Experimental Setup}

\paraf{Testbed.}
We evaluate \sysname on a 16-server cluster with 128 NVIDIA H800 GPUs.
Each server contains eight GPUs, 160 Intel Xeon CPU cores, and 1.8 TB of host memory.
The software stack uses PyTorch 2.8.0 and CUDA 12.2 with driver 535.161.08.

\parabf{Models \& Datasets.}
We evaluate \sysname on two representative MLLM training workloads: MLLM-Understanding and MLLM-Generation.
For MLLM-Understanding, we use Qwen3-30B-A3B~\citep{qwen2025qwen3} as the backbone LLM and a 1.3B-parameter ViT~\citep{dosovitskiy2021image} as the vision encoder.
The model is trained on image-text pairs from a large-scale web-crawled dataset to learn visual-textual understanding and reasoning.
For MLLM-Generation, we use the same backbone LLM and vision encoder, and attach a 20B-parameter MMDiT~\citep{esser2024scaling} as the multimodal generator.
This follows the model architecture in Qwen-Image~\citep{wu2025qwen}.
The model is trained on internal text-to-image (T2I), text-image-to-image (TI2I), and image-to-image (I2I) pairs to generate images from textual and visual inputs.
Both workloads use an 8K input sequence length for the LLM backbone.
All modules remain trainable, with gradients enabled throughout training.

\parabf{Baselines \& Metrics.}
We compare \sysname with two representative MLLM training baselines.
The first is Optimus~\citep{feng2025optimus}, a compute-efficient design that runs encoder forward, LLM forward and backward, and generator backward sequentially.
The second is Megatron-DistTrain~\citep{zhang2025disttrain}, a memory-efficient design that integrates the encoder and generator as the first and last stages of the pipeline.
We call it Megatron-DistTrain since we use Megatron's open-source implementation of DistTrain.
These two baseline systems are widely adopted in production MLLM training~\citep{team2025longcat,huang2025step}.
We report training iteration time and peak GPU memory usage, the two primary metrics for MLLM training efficiency and scalability.

\begin{figure*}[t!]
    \centering
    \includegraphics[width=0.86\linewidth]{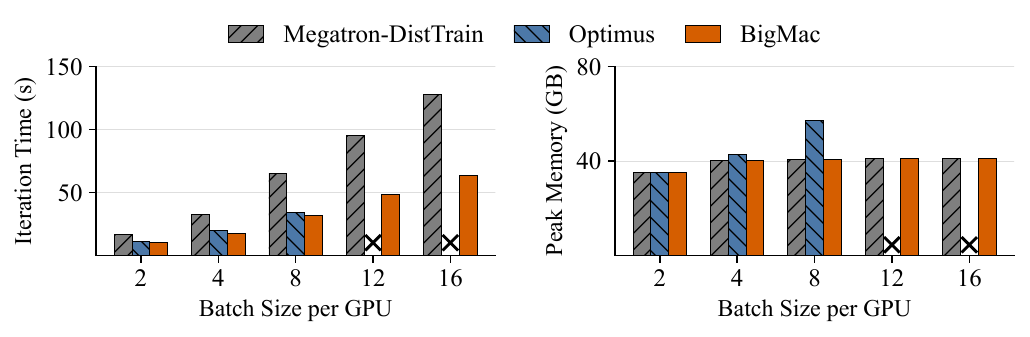}
    \vspace{-0.15in}
    \caption{Overall performance of \sysname under training MLLM-Understanding.}
    \vspace{-0.17in}
    \label{fig:eval:overall:vit}
\end{figure*}

\begin{figure*}[t!]
    \centering
    \includegraphics[width=0.86\linewidth]{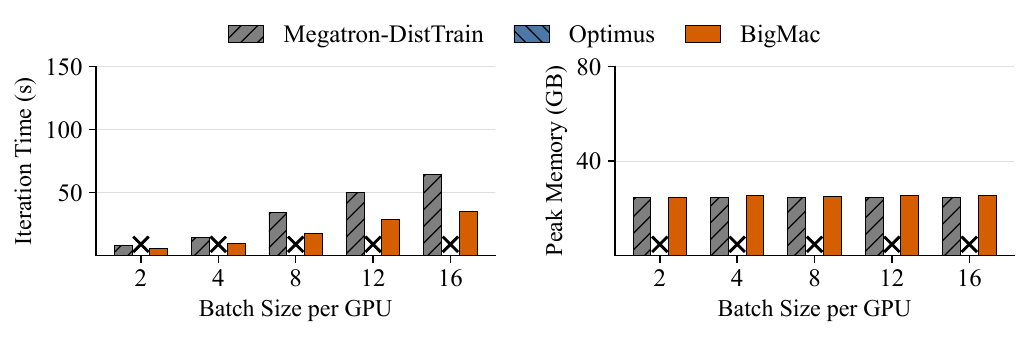}
    \vspace{-0.15in}
    \caption{Overall performance of \sysname under training MLLM-Generation.}
    \vspace{-0.17in}
    \label{fig:eval:overall:dit}
\end{figure*}

\subsection{Overall Performance}
\paraf{MLLM-Understanding.}
We first evaluate \sysname on the MLLM-Understanding workload.
Figure~\ref{fig:eval:overall:vit} reports iteration time and peak GPU memory usage under different per-GPU batch sizes.
\sysname consistently achieves the lowest iteration time, improving over Optimus by 1.08$\times$--1.1$\times$ and over Megatron-DistTrain by 1.6$\times$--1.9$\times$.
Compared with Megatron-DistTrain, \sysname removes the execution interference between the encoder and LLM, thereby reducing pipeline bubbles.
Compared with Optimus, \sysname keeps activation memory low, reducing memory pressure and avoiding the expensive memory operations.
For peak memory, \sysname and Megatron-DistTrain maintain a stable memory usage when batch size increases, whereas Optimus's memory usage grows rapidly because it retains encoder activations across the LLM pipeline.
When the per-GPU batch size exceeds 8, Optimus runs out of memory; cross markers denote these OOM points.
Overall, \sysname achieves the best of both compute and memory efficiency.

\parabf{MLLM-Generation.}
We then evaluate \sysname on the MLLM-Generation workload as shown in Figure~\ref{fig:eval:overall:dit}.
Optimus runs out of memory at all batch sizes since it needs to retain all LLM activations until the generator finishes.
Compare with Megatron-DistTrain, \sysname reduces iteration time by 1.5$\times$--1.9$\times$ by eliminating the execution interference between the encoder, LLM, and generator.
For peak memory, \sysname and Megatron-DistTrain maintain a stable memory usage when batch size increases.

\begin{figure*}[t!]
    \centering
    \includegraphics[width=0.86\linewidth]{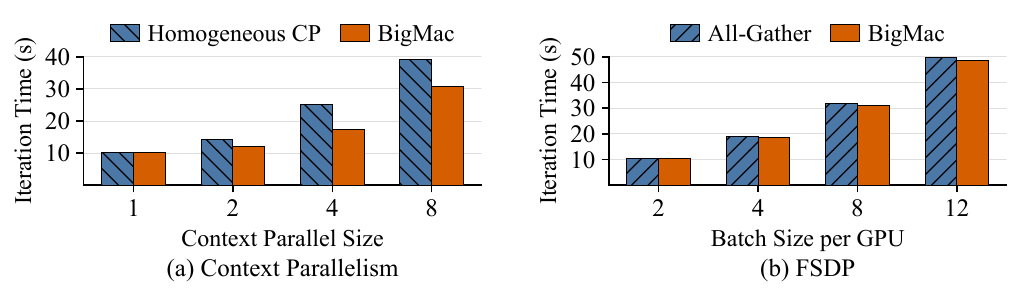}
    \vspace{-0.15in}
    \caption{Case study of \sysname's compatibility.}
    \vspace{-0.17in}
    \label{fig:eval:case}
\end{figure*}

\subsection{Case Study}

\paraf{Context parallelism.}
We evaluate whether \sysname can benefit from decoupling context parallelism (CP) across MLLM components.
The homogeneous CP baseline applies the same CP degree to both the encoder and the LLM backbone, which is suboptimal because the two modules have different model architectures and process different tokens.
\sysname instead allows the encoder and LLM to use separate CP groups.
Figure~\ref{fig:eval:case}(a) compares \sysname with homogeneous CP as the LLM CP degree increases.
\sysname reduces iteration time by up to 1.45$\times$, demonstrating the effectiveness of decoupled CP for MLLM training.

\parabf{FSDP.}
We next evaluate \sysname's compatibility with FSDP-trained modality modules.
The baseline uses standard FSDP \texttt{all-gather} communication to materialize full parameters before each FSDP-wrapped encoder or generator operator.
\sysname replaces this collective synchronization with one-sided pull, allowing each rank to proceed according to its local pipeline schedule.
This reduces training iteration time by up to 1.03$\times$ while preserving FSDP's sharded-parameter memory benefit.

\parabf{Bitwise alignment.}
\sysname changes only the order in which module gradients are accumulated, while preserving the synchronous training semantics of the original pipeline.
To isolate pipeline effects from kernel-level nondeterminism, we disable FlashAttention and use deterministic compute operators in this study.
We then run \sysname and Megatron-LM on the same data stream under different per-GPU batch sizes and record their training losses.
Across the first 128 iterations, the average absolute loss difference between the two systems remains below $10^{-3}$, indicating that \sysname preserves
training semantics.

\section{Discussion and Future Work}
\label{sec:future}

MegaScale-Omni~\citep{xue2026megascale} follows a similar design philosophy for training MLLM-understanding models while reducing encoder memory usage.
However, unlike \sysname, it targets encoder-only MLLMs and does not support modality generators.
It also directly patches Megatron-LM's LLM pipeline, limiting scheduling flexibility and making it difficult to integrate with different training frameworks.
Moreover, its simplified pipeline design lacks support for production LLM training optimizations such as point-to-point communication/computation overlap and combined 1F1B.

There are two important directions for future work.
First, \sysname currently focuses on preserving a user-specified
LLM pipeline schedule while nesting the encoder and generator units into it.
Extending this design to a broader class of LLM pipeline schedules,
such as zero-bubble pipeline parallelism~\citep{qi2023zero}, would further improve its
applicability to emerging large-scale training systems. Second, although
\sysname removes cross-module execution interference, multimodal data
heterogeneity can still create load imbalance within the encoder and generator itself,
since different samples may induce different encoder costs.
Future work can explore data-aware packing to reduce this intra-module
imbalance.

\section{Conclusion}
\label{sec:conclusion}
\sysname addresses the compute--memory tradeoff in MLLM training with a dependency-safe nested pipeline.
It preserves the user-specified LLM schedule and nests encoder/generator units only when their dependencies are satisfied, eliminating cross-module interference while keeping modality encoder and generator's activation memory $O(1)$.
Across MLLM-understanding and generation workloads, \sysname achieves 1.08$\times$--1.9$\times$ speedups while maintaining stable memory usage as batch size increases.
\label{lastpage}

{
\bibliographystyle{iclr2026_conference}
\bibliography{xin}}

\clearpage

\end{document}